\documentclass{article}
\usepackage{amsmath,graphicx,mlspconf,xcolor,soul}
\usepackage{amsfonts}
\usepackage{caption}
\usepackage{subcaption}
\usepackage{url}
\usepackage{algorithm,algpseudocode}
\algnewcommand{\LineComment}[1]{\State \(\triangleright\) #1}

\title{Layer Ensembles}
\name{Illia Oleksiienko and Alexandros Iosifidis\thanks{This work has received funding from the European Union’s Horizon 2020 research and innovation programme (grant agreement No 871449 (OpenDR)).}}
\address{Department of Electrical and Computer Engineering, Aarhus University, Denmark}

\begin{document}
%\ninept

\maketitle

\begin{abstract}
Deep Ensembles, as a type of Bayesian Neural Networks, can be used to estimate uncertainty on the prediction of multiple neural networks by collecting votes from each network and computing the difference in those predictions. In this paper, we introduce a method for uncertainty estimation which considers a set of independent categorical distributions for each layer of the network, giving many more possible samples with overlapped layers than in the regular Deep Ensembles. We further introduce an optimized inference procedure that reuses common layer outputs, achieving up to 19x speed up and reducing memory usage quadratically. We also show that the method can be further improved by ranking samples, resulting in models that require less memory and time to run while achieving higher uncertainty quality than Deep Ensembles.
\end{abstract}

\section{Introduction}
\label{sec:introduction}

Uncertainty estimation in neural networks is an important task for critical problems, such as autonomous driving, medical image analysis, or other problems where silent failures of machine learning systems can lead to high-cost damages or endanger lives. 
Bayesian Neural Networks (BNNs) \cite{1995bayessian_nn, wilson2020bayesian, charnock2020bayesian,magris2022bayes_survey} provide a tool to estimate prediction uncertainty by exploiting a distribution over the network weights and sampling a set of models with slightly different predictions for a given input. This difference in the predictions expresses the uncertainty of the network, while the mean of all predictions is used as the prediction of the network. 
The selection of the adopted distribution affects the computational requirements and statistical quality of the network, with Gaussian distribution resulting in Bayes By Backpropagation (BBB) \cite{blundell2015weight} and Hypermodel \cite{dwaracherla2020hypermodels} methods, Bernoulli distribution in Monte Carlo Dropout (MCD) \cite{2016dropout}, and Categorical distribution in Deep Ensembles \cite{osband2018randomized}.

We introduce Layer Ensembles, which consider a set of weight options for each layer that are sampled using independent Categorical distributions, resulting in a high number of models that can have common layer samples.
%In the uncertainty quality experiments, w
We show that Layer Ensembles achieve better uncertainty quality than Deep Ensembles for the same number of parameters, and they allow to dynamically change the number of samples to keep the best ratio between the uncertainty quality and time cost.

The proposed method is a good fit for applications where the computational budget can fluctuate, and the requirements for model speed or uncertainty quality are flexible. Such scenarios include autonomous driving, robot control and Machine Learning on personal devices. In the case of a high computational budget, the selected sample set can be of a bigger size, leading to a better uncertainty estimation process, while if there are not enough device capabilities, we can sacrifice some of the uncertainty quality to keep the system operation at the tolerable speed. This process is explained in details in Section \ref{sec:lens}.

\section{Related Work}\label{sec:related_works}
Output uncertainty estimation in deep neural networks is usually done by approximating expectation and covariance of outputs using the Monte Carlo integration with a limited number of weight samples. This can be simplified to performing inference a few times using different randomly sampled weights for the network, and then computing the mean and variance of the network output vectors. 
Epistemic Neural Networks (ENNs) \cite{osband2021epistemic} propose a framework to estimate an uncertainty quality of a model by generating a synthetic dataset and training a Neural Network Gaussian Process (NNGP) \cite{lee2018nngp} on it that represents a true predictive distribution. 
The model of interest is then evaluated by the KL-divergence \cite{1951_kl_information} between the true predictive distribution from NNGP and the predictive distribution of the model of interest. 

Monte Carlo Dropout (MCD) \cite{2016dropout}, instead of only using Dropout \cite{srivastava2014dropout} layers as a form of regularization during training to avoid overtrusting particular neurons, keeps the Dropout layers also during inference. This has the effect of adopting a Bernoulli distribution of weights and sampling different models from this distribution. %It achieves the worst uncertainty quality in ENNs experiments. 
Bayes By Backpropagation (BBB) \cite{blundell2015weight} considers a Gaussian distribution over network weights, which is estimated using the reparametrization trick \cite{kingma2014autoencoding} that allows to use regular gradient computation. %It outperforms MCD, but has worse results than the methods below.

Variational Neural Networks (VNNs) \cite{oleksiienko2022vnn} can be considered in the same group as MCD and BBB from the Bayesian Model Averaging perspective, where sampled models can lie in the same loss-basin and be similar, i.e., describing the problem from the same point of view, as explained in \cite{wilson2020bayesian}. 
VNNs consider a Gaussian distribution over each layer, which is parameterized by the outputs of the corresponding sub-layers. 
Hypermodels \cite{dwaracherla2020hypermodels} consider an additional hypermodel $\theta = g_\nu(z)$ to generate parameters of a base model $f_\theta(x)$ using a random variable $z \sim \mathcal{N}(0, I)$ as an input to the hypermodel. 

Deep Ensembles \cite{osband2018randomized} have a better uncertainty quality than all other discussed methods and can be viewed as a BNN with a Categorical distribution over weights, with the ideal number of weight samples being equal to the number of networks in the ensemble. The addition of prior untrained models to Deep Ensembles, as described in \cite{osband2018randomized}, improves the uncertainty quality of the network. 
Deep Sub-Ensembles \cite{valdenegro2019subens} split the neural network into two parts, where the first part contains only a single trunk network, and the second part is a regular Deep Ensemble network operating on features generated by the trunk network.
This reduces the memory and computational load compared to the Deep Ensembles and provides a trade-off between the uncertainty quality and resource requirements. 
Batch Ensembles \cite{wen2020batchens} optimize Deep Ensembles by using all weights in a single matrix operation and using Hadamard product instead of matrix multiplication that increases inference speed and reduces memory usage. 

Kushibar et al. \cite{kushibar2022medicalle} propose to use a single deterministic network to generate different outputs by using multiple early-exits branches \cite{panda2016ealyexit} from the same network, and compute the variance in those outputs. 
%The idea of combining information from multiple early exits has been investigated before, e.g., \cite{passalis2020hierearlyexits}, but instead of computing the variance of such outputs the method combines them using Bag-of-Features and improve the point predictions.
Contrary to this approach using a single deterministic network, we propose Layer Ensembles in the context of Bayesian Neural Networks by considering an independent Categorical distribution over weights of each layer, as described in the following section.

\section{Layer Ensembles}
\label{sec:lens}
In this section, we first provide a mathematical definition of Layer Ensembles structure. Then, we provide the intuition behind Layer Ensembles and the relations between Layer Ensembles and other network structures.
Furthermore, we show how inference of Layer Ensembles can be optimized by reusing common layer outputs. Finally, we propose a method for selecting the best sample combinations based on the quality of uncertainty metric. This metric is computed using Epistemic Neural Networks (ENNs) \cite{osband2021epistemic} experiments with a synthetic dataset that has ground-truth uncertainty values, as described in details in Section \ref{sec:sample_ranking}.

We consider a neural network $F(x, w)$ with $N$ layers which takes $x$ as input and is parameterized by the weights $w$.
A Deep Ensemble network is formed by $K$ identically structured networks, each formed by $N$ layers and the weights of each network, i.e., $w_i, \:i \in [1, K]$, are trained independently.
We formulate Layer Ensembles as a stochastic neural network $F(x, w)$ with $N$ layers $\mathrm{LE}_i(x, w_q^i), \:i \in [1, K], \:q \in [1, N]$ and $K$ weight options for each layer:
\begin{align}
\begin{split}
    &w_q^i \sim \mathrm{Categorical}(K). \\
\end{split}
\end{align}
This results in the same memory structure as for Deep Ensembles, with $KN$ weight sets for an ensemble of $K$ networks each formed by $N$ layers.
However, Layer Ensembles allow for connections between the layers of different weight sets, by sampling different layer options to form a network in the ensemble. This greatly increases the number of possible weight samples, while those can contain identical subnetworks. This can be used to speed up the inference of a set of sampled layers.

\begin{figure}
     \centering
     \begin{subfigure}[b]{0.47\linewidth}
         \centering
         \includegraphics[width=\linewidth]{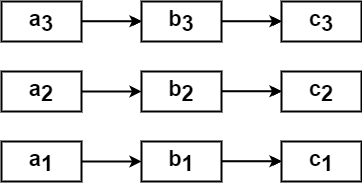}
         \caption{Deep Ensemble structure}
         \label{fig:ensemble_connections:de}
     \end{subfigure} \:\:
     \begin{subfigure}[b]{0.47\linewidth}
         \centering
         \includegraphics[width=\linewidth]{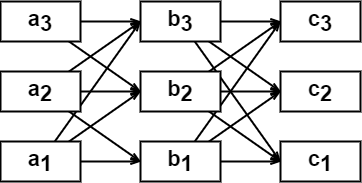}
         \caption{Layer Ensemble structure}
         \label{fig:ensemble_connections:le}
     \end{subfigure}
     \begin{subfigure}[b]{\linewidth}
         \centering
         \includegraphics[width=0.49\linewidth]{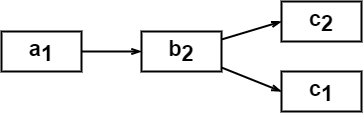}
         \caption{Two samples of a Layer Ensemble network with common first two layer options}
         \label{fig:ensemble_connections:samples}
     \end{subfigure}
    \caption{Example structures of (a) Deep Ensembles and (b) Layer Ensembles for a 3-layer network with 3 ensembles ($N=3$, $K=3$). While the memory structure remains identical, Layer Ensembles have many more options for sampling that can be optimized considering the common layers in samples. Layer Ensembles with common layers earlier in the architecture lead to faster inference (c).}
    \label{fig:ensemble_connections}
\end{figure}

The intuition behind Layer Ensembles is to define an ensemble for each layer. These ensembles are independent and can have a different number of members in each ensemble, leading to a highly flexible network structure. If a single random variable controls all layer-wise ensembles, and they have an identical number of members, this leads to the well-known Deep Ensemble neural network.
Fig.~\ref{fig:ensemble_connections} illustrates how the same memory structure of the ensembles is used in Deep Ensembles (Fig.~\ref{fig:ensemble_connections:de}) and Layer Ensembles (Fig.~\ref{fig:ensemble_connections:le}).

Training of Layer Ensembles is done by using a regular loss function and averaging over the outputs of different network samples.
The number of network samples for Deep Ensembles is usually equal to the number of networks $K$, meaning that all the ensembles are used for inference.
The same strategy is not required for Layer Ensembles, as each layer option can be included in multiple networks. This means that one can select a few network samples per inference and expect that, with the sufficient amount of training steps, all the layer weights will be trained.

Following \cite{osband2018randomized}, we consider an output of a prior untrained Layer Ensemble network added to the output of the trained network, using the same draws from the random distributions for both untrained and trained networks.
Experiments show that the addition of prior networks improves the uncertainty quality of Layer Ensembles by a factor of $2$ for each number of ensembles that was tested.

The freedom of connections in Layer Ensembles is high, and by applying different restrictions to the number of weight options for different layers, as well as to the random variables that control these ensembles, we can define Deep Ensembles \cite{osband2018randomized} and Deep Sub-Ensembles \cite{valdenegro2019subens} as special cases of Layer Ensembles.
Considering a Layer Ensemble network with $N$ layers and $K$ weight options for each layer, we can sample $K^N$ possible models.
Sampling $K$ networks with none of the layer options used in two different networks corresponds to a Deep Ensemble network.
By selecting different number of ensembles per layer, we can achieve Deep Sub-Ensembles by using one ensemble for the first $T$ layers and $K$ ensembles for the remaining $N-T$ layers, resulting in a single trunk network and in an ensembled tail network.
Experimenting with the number of ensembles for each layer can result in interesting new methods for specific analysis problems and is a direction for future work.

\subsection{Inference Optimization}
Layer Ensembles can reuse outputs of identical subnetworks processing the input $x$ when they are used in different networks.
Fig.~\ref{fig:ensemble_connections:samples} shows an example of two Layer Ensembles where the first two layers are identical, and only the last layer has different weights.
Instead of computing $c_2(b_2(a_1(x)))$ and $c_1(b_2(a_1(x)))$ independently, one can compute the common layer $V = b_2(a_1(x))$ first, and then $c_2(V)$ and $c_1(V)$.

To achieve this, we have to consider a set of samples together, instead of applying each network sample independently, and identify layer outputs that can be reused for a given input.
Only when the first $G$ layers of two sampled networks are the same, their outputs are identical. For example, in Fig. \ref{fig:ensemble_connections:samples}, the two sampled networks are $(a_1,b_2,c_1)$ and $(a_1, b_2, c_2)$, which have $G=2$ first identical layers, outputs of which should be computed only once and reused for the second sample.
Considering more than two samples, the possible overlapping layers have a hierarchical structure, where a layer sample $l_1$ for the first layer can be reused by some set of networks, but for the next layer this set will be split up between the possible samples of the second layer, creating a tree-like structure of reusable layer outputs.
We can traverse this structure and compute all network outputs simultaneously by using a depth-first search on this graph of overlapping layers.

\begin{algorithm}
  \caption{Optimized Layer Ensembles}
  \label{alg:ole}
  \begin{algorithmic}[1]
    \Require{Network $F(x)$, list of sorted samples $S$, layer index $i$, input $x$}
    \Function{OLE}{$F, S_i, i, x$}
      \State $\mathrm{result} \gets []$
      \State $s_{i+1} \gets []$
      
      \If{$i = \mathrm{size}(s)$}
        \Return $[x]$ \Comment{Final layer computed}
      \EndIf
      
      \State $s_l \gets S_i[0]$
      \State $l \gets F[i][s_l](x)$ \Comment{First sampled option for layer $i$}
      
      \For{$t \in [0 .. \mathrm{size}(S_i)]$} \Comment{For each sample}
        
        \If{$S_i[t] \neq s_l$}
            \State $\mathrm{result} = \mathrm{result} \cup \mathrm{OLE}(F, s_{i+1}, i+1, l)$
            \State $s_l \gets S_i[t]$
            \LineComment{Next sampled option for layer $i$}
            \State $l \gets F[i][s_l](x)$ 
            \State $s_{i+1} \gets []$
        \EndIf
        
        \LineComment{Update sub-samples list for input $l$}
        \State $s_{i+1} \gets s_{i+1} \cup \mathrm{tail}(s_i[t])$

      \EndFor
      
      \State $\mathrm{result} = \mathrm{result} \cup \mathrm{OLE}(F, s_{i+1}, i+1, l)$ %\Comment{Compute for final $s_l$}
      
      \State \textbf{return} result
    \EndFunction
    
    \State \textbf{return} OLE$(F, S, 0, x)$
  \end{algorithmic}
\end{algorithm}

Algorithm \ref{alg:ole} implements an Optimized Layer Ensembles (OLE) function that recursively computes the output of a Layer Ensemble network for a set of sorted layer samples.
Layer samples are represented as a set of selected options for each layer, such as $[1, 2, 2]$ and $[1, 2, 1]$ for the models in Fig.~\ref{fig:ensemble_connections:samples}.
These samples are sorted in ascending order by the first-most values, while using later indices in case of identical previous values.
This allows to have the most overlapping samples in a sequence, giving the possibility to optimize layer executions, as a layer option should be called only once for the same input and used by all samples that share it.
After the current layer option is used, there is no need to keep its output in memory anymore, as it will never be used later. This improves the memory requirement of the model.
The results of the OLE function is an array of outputs for all runs using this layer, which means that in order to run a full set of samples the OLE function needs to be called with network function $F$, samples list $S$, layer index $i=0$, and the input to the network $x$.

\begin{figure}
\centering
    \includegraphics[width=1\linewidth]{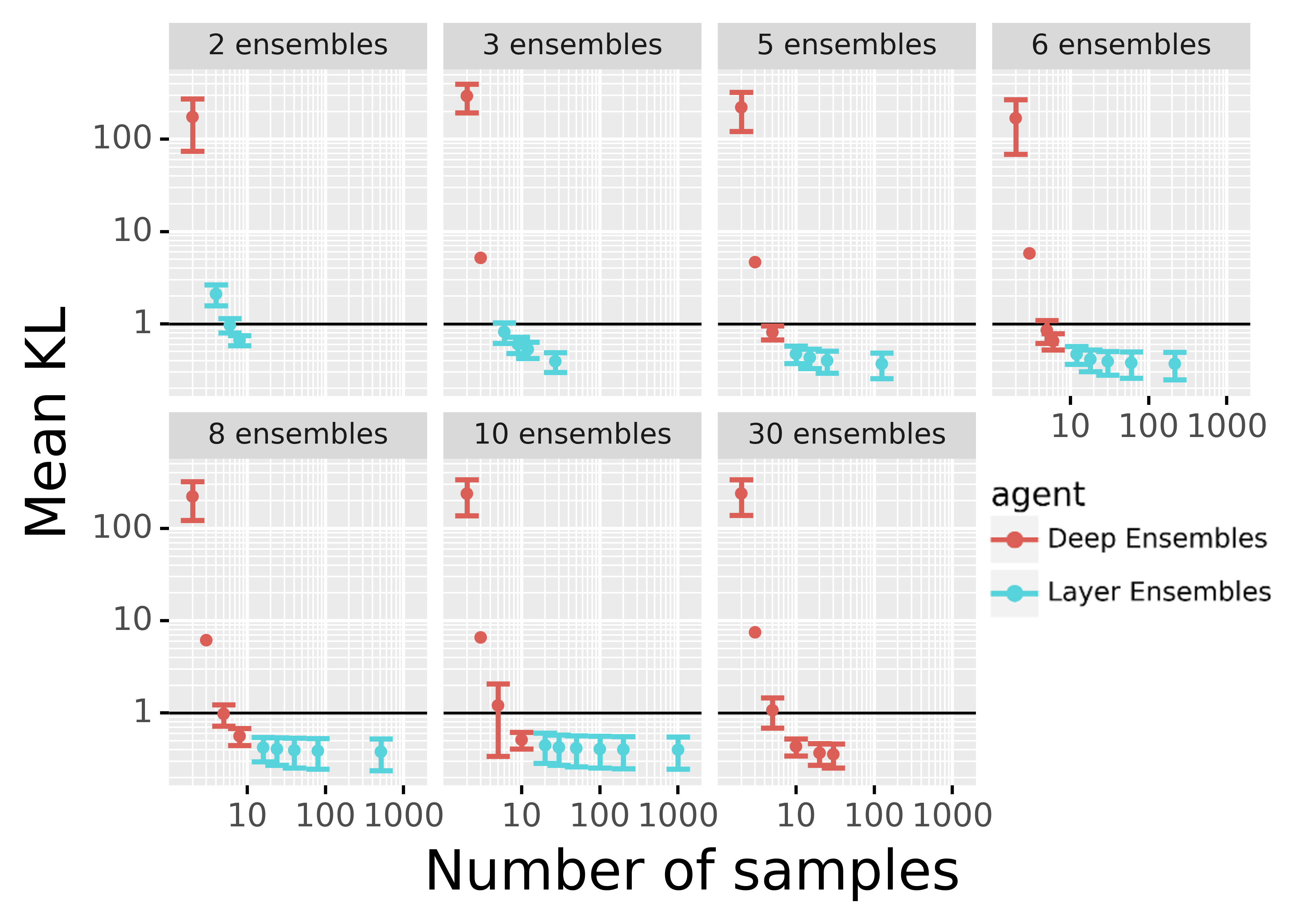}
    \caption{Comparison of mean KL values with 1 STD range for Deep Ensembles and Layer Ensembles with random unique layer samples, averaged across all experiment parameters.}
    \label{fig:results-kl-all}
\end{figure}

\subsection{Layer sample ranking}\label{sec:sample_ranking}
Following Optimized Layer Ensembles, we can improve the speed of the model and the quality of uncertainty by selecting which of the sampled networks should be used during inference, since networks that share most of the layer options can be too similar from the Bayesian Model Averaging perspective.

One way to decrease the computational cost is to randomly select fewer layer samples, resulting in slightly lower quality of uncertainty, as shown in Fig.~\ref{fig:results-kl-all}.
Another option is to rank layer samples based on the uncertainty quality on the validation set and use the best layer samples when using a particular number of samples.

We follow the Epistemic Neural Networks (ENNs) \cite{osband2021epistemic} framework to estimate the uncertainty quality of the method and use it to select the best layer samples, and to compare the proposed method with the state-of-the-art uncertainty estimation approaches.
ENNs consider a regression task $y = f(x) + \epsilon$ and generate a synthetic dataset $D_T = \{(x, y)_t$ for $t \in [0, T-1]\}$, where $x$ is a $D_x$-dimensional input vector, $y$ is an output scalar, $\epsilon \sim \mathcal{N}(0, \sigma^2)$ is a random noise, and $T=D_x \lambda$ is a dataset size.
A Neural Network Gaussian Process (NNGP) \cite{lee2018nngp} is trained on the dataset to represent the true model of the data.
For each data point, a model of interest should provide a prediction $\mu$ and an uncertainty in that prediction $\sigma^2$, which is modeled by a one-dimensional Gaussian distribution $\mathcal{N}(\mu, \sigma^2)$.
Given the predictions from both the true NNGP model and a model of interest $M$, the uncertainty quality score $Q(M)$ is computed as:
\begin{align}
\begin{split}
    &Q(M) = \frac{1}{T}\sum^{T-1}_{t=0} \mathrm{KL}(\mathcal{N}_M \parallel \mathcal{N}_{\mathrm{NNGP}}), \\
    & \mathcal{N}_M = \mathcal{N}(\mathbb{E}[M(x_t)], \mathrm{Var}[M(x_t)]), \\
    & \mathcal{N}_{\mathrm{NNGP}} = \mathcal{N}(\mathbb{E}[\mathrm{NNGP}(x_t)], \mathrm{Var}[\mathrm{NNGP}(x_t)]), \label{eq:uncertainty}
\end{split}
\end{align}
where $M$ and $\mathrm{NNGP}$ are the model of interest and the true NNGP model, respectively, and $\mathrm{KL}$ is a Kullback–Leibler divergence function \cite{1951_kl_information}.
As can be seen in Eq. (\ref{eq:uncertainty}), the quality of uncertainty is defined based on how close the model's output and its uncertainty, represented by a Gaussian distribution, to the distribution of the ground-truth model.
This means that in cases where the network is not competent to provide an accurate output, it should be uncertain, and when the network is competent (e.g., when it operates in a well-explored area), it should be certain in its output. %re is not enough information to make a competent prediction, the model should be uncertain, and if an input is in a well-explored area, the model should be certain in its predictions.

\begin{figure}
\centering
    \includegraphics[width=1\linewidth]{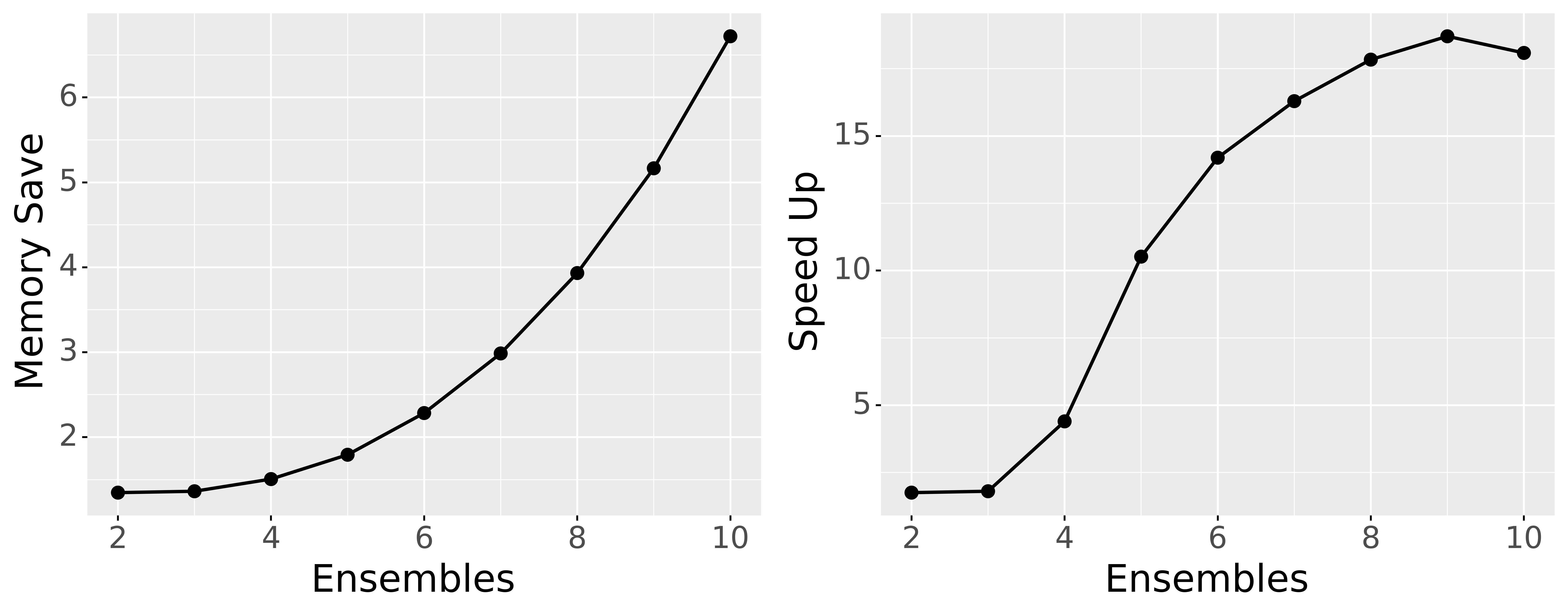}
    \caption{Speed up and memory saved during inference of Optimized Layer Ensembles, compared to the regular Layer Ensembles for different number of ensembles of a 4-layer CNN for MNIST classification. This excludes memory used for the ML framework and model weights.}
    \label{fig:ole}
\end{figure}

\begin{figure*}
\centering
    \includegraphics[width=1\linewidth]{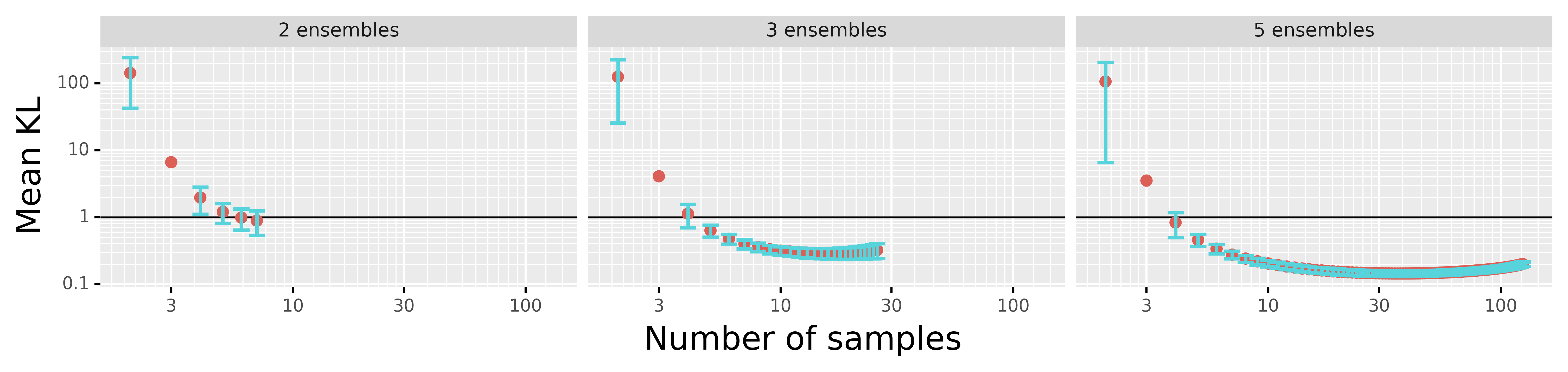}
    \caption{Comparison of mean KL values with 1 STD range for Layer Ensembles with different number of ensembles and sampled layers, averaged across all experiment parameters. The best layer samples are selected based on the validation set and evaluated on the test set.}
    \label{fig:ranking}
\end{figure*}

Let us consider the full set of layer samples $S^J = \{s_j | j \in [1, J]\}$, where $J$ is a number of all combinations that can be computed by multiplying all layer-wise numbers of ensembles. To reduce the computational load of layer sample ranking, we introduce an iterative process of selecting the best layer samples by starting from a single layer sample with the best mean error $s^1$:
\begin{equation}
    s^1 = \underset{s_j}{\mathrm{argmax}} \:\: Q(M^{\{s_j\}}),
\end{equation}
where $Q(\cdot)$ is the uncertainty quality score function, $M^{\{s_j\}}$ is the Layer Ensemble model applied to a set of layer samples, containing a single sample $s_j$.
Given a set of optimal layer samples $S^P = \{s^P_i | i \in [1..P] \}$ of size $P$, the next layer sample set is created by finding the best addition to the already existing set:
\begin{align}
\begin{split}
    s^{P+1} &= \underset{s_j}{\mathrm{argmax}} \:\: Q(M^{S^P \cup s_j}), \\
    S^{P+1} &= S^P \cup s^{P+1}.
\end{split}
\end{align}

Following the above process, we select the best layer samples set $S^P$ for each number of samples $P$ from $1$ to $J$ based on the performance on the validation dataset. The selected samples are then used to perform inference during testing. This means that for a given number of samples $P$, a deterministic ensemble model, which is expected to provide the highest uncertainty quality, is used during inference. This provides flexibility in the inference phase as, based on the computational budget available for performing inference on different input data samples, one can choose an appropriate value for $P$. 
%The processing system may have different computation budget at different input frames, which can be utilized by creating a lookup table with the desired number of samples $P$ as keys and the corresponding layer samples sets $S^P$ as values.
When the available computational budget is low, a lower number of samples $P$ can be selected and the corresponding layer samples set $S^P$ is used to perform inference, reducing the computational requirement and providing high-quality uncertainty for the current model. In the case of an increased computational budget being available, a higher number of samples $P$ can be selected to improve the quality of the uncertainty.

\section{Experiments}\label{sec:uncertainty_exepriments}

We implement Layer Ensembles inside the ENNs JAX repository \cite{github_enn} and implement the experiments using the following parameters: $D_x \in \{10, 100, 1000\}$, $\lambda \in \{1, 10, 100\}$, and $\epsilon \in \{0.01, 0.1, 1\}$. The experiments are repeated with all combinations of $(D_x, \lambda, \epsilon)$ parameters and with $10$ different random seeds. The results are then averaged, computing the mean and variance of uncertainty quality scores for each method.
Fig.~\ref{fig:results-kl-all} provides a comparison between Deep Ensembles and Layer Ensembles for different number of ensembles and number of samples. Layer Ensembles start to achieve good uncertainty quality with only 3 ensembles and outperform Deep Ensembles for the same number of ensembles used. This means that the memory footprint is much lower for Layer Ensembles.
The effect of using OLE, compared to the original Layer Ensembles, on the time and memory consumption is presented in Fig. \ref{fig:ole}.
The comparison is performed for a full set of samples for different number of ensembles, showing that the memory usage is reduced quadratically, and the inference speed is up to $19$ times higher for OLE.
Since Machine Learning frameworks usually use a data-flow approach to computations, the Optimized Layer Ensembles benefit from these asynchronous computations, as the only data gathering across different Layer Ensemble samples is performed at the end of the inference function. However, there is a limit on how much the computations can be parallelized, which reduces the speed-up gains when increasing the number of ensembles. This limit depends on the number of layers and the computational capabilities of the physical device. In the case of a 4-layer CNN and a 2080Ti GPU, this limit is reached at 9 ensembles.

\begin{figure}
\centering
    \includegraphics[width=0.95\linewidth]{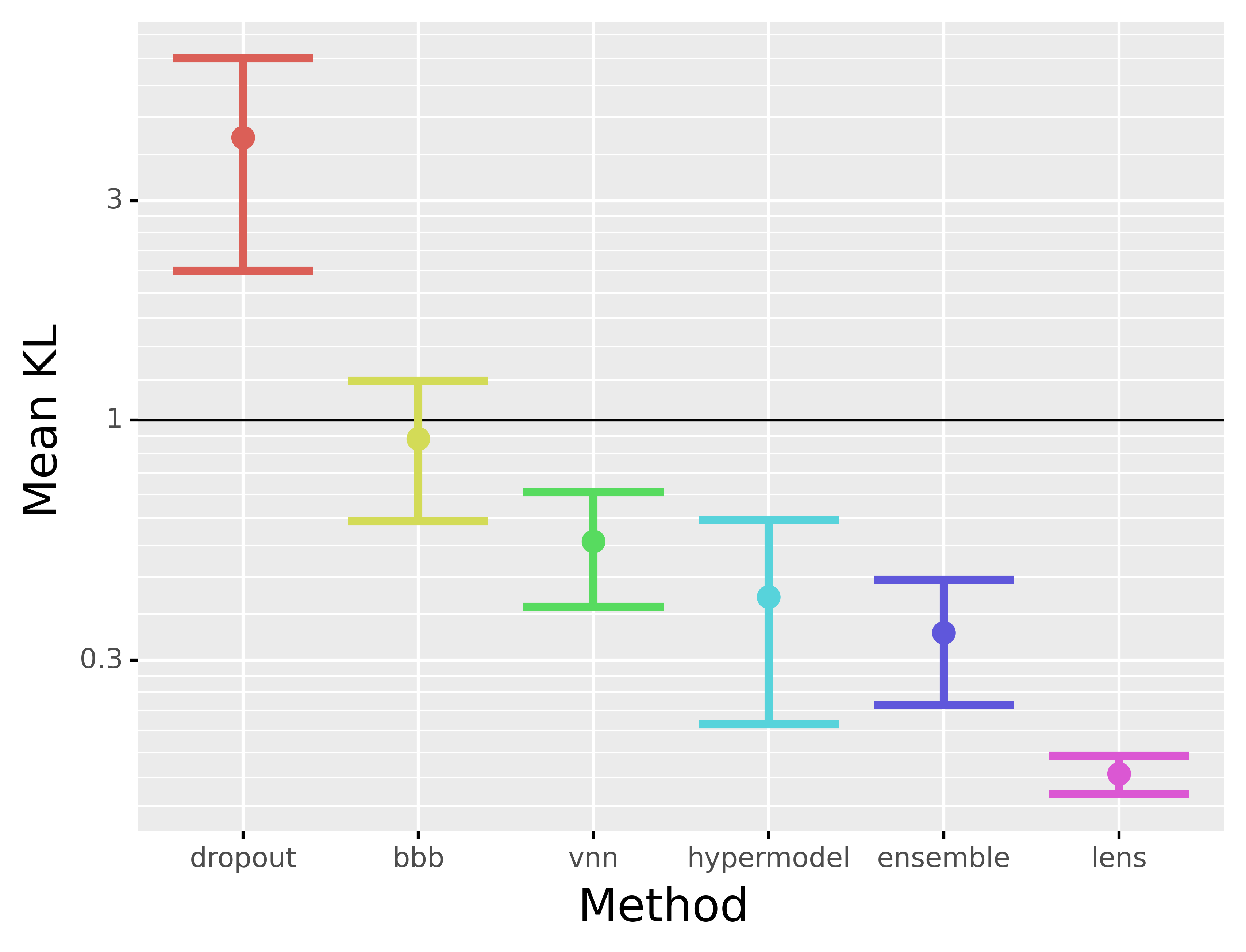}
    \caption{Comparison of mean KL values with 1 STD range for BNNs and the best Layer Ensembles model with $5$ ensembles using sample ranking.}
    \label{fig:lens-vs-bnn}
\end{figure}

Fig.~\ref{fig:ranking} illustrates the uncertainty quality results of each best layer sample set of Layer Ensembles.
With a number of ensembles higher than 2, the optimal uncertainty quality is increased up to a certain number of layer samples and starts decreasing after that. This means that it is beneficial for both inference speed and uncertainty quality to not use all the available layer samples. 
Layer Ensembles with $5$ ensembles achieves the best uncertainty quality at $36$ layer samples, which is much lower than the number of $125$ possible layer samples.
Even at $20$ layer samples, the uncertainty quality is just $6\%$ lower, but it is still $2$ times better than the Deep Ensembles with 30 ensembles, while it uses $6$ times less memory, and it is at least $1.5\times$ faster, as the speed can be improved by Optimized Layer Ensembles inference procedure but depends on the overlap between layer samples.

We compare the best Layer Ensembles model for $5$ ensembles, created using the sample ranking process, to the state-of-the-art BNN methods following the earlier described ENNs framework.
The evaluation results for MCD \cite{2016dropout} (dropout), BBB \cite{blundell2015weight}, VNNs \cite{oleksiienko2022vnn}, Hypermodels \cite{dwaracherla2020hypermodels}, Deep Ensembles \cite{osband2018randomized} (ensemble) and the proposed Layer Ensembles (lens) are shown in Fig. \ref{fig:lens-vs-bnn}.
Layer Ensembles achieve the highest quality of uncertainty among the tested methods, with the lowest variance of the uncertainty quality score across the experiments.

\section{Conclusions}\label{sec:conclusions}
In this paper, we proposed a novel uncertainty estimation method called Layer Ensembles, which corresponds to a Bayesian Neural Network with independent Categorical distribution over weights of each layer.
We showed that Layer Ensembles use parameters more effectively than Deep Ensembles and provide a flexible way to balance between inference time and model uncertainty quality.
We showed that the inference of Layer Ensembles can be optimized by performing the same computations once, which increases the inference speed by up to $19$ times and reduces memory usage quadratically.
Finally, we proposed a layer sample ranking system that allows to select the best layer samples based on the combined uncertainty quality, leading to a high increase in uncertainty quality and reducing both memory and time requirements.

\bibliographystyle{IEEEbib}
\bibliography{bibliography}

\end{document}